\documentclass[11pt]{article}

\usepackage[preprint]{acl}

\usepackage{times}
\usepackage{latexsym}

\usepackage[T1]{fontenc}

\usepackage[utf8]{inputenc}

\usepackage{microtype}
\usepackage{inconsolata}
\usepackage{graphicx}
\usepackage{booktabs}
\usepackage{pifont}
\usepackage{listings}
\usepackage{multirow}
\usepackage{adjustbox}
\usepackage{amsmath}
\usepackage{amssymb}
\usepackage{xcolor}
\usepackage{tabularx}
\usepackage{float}
\usepackage{tabularx}
\usepackage[most]{tcolorbox}
\usepackage{listings}
\tcbuselibrary{listings,breakable,skins}
\usepackage{graphicx}
\usepackage{subcaption}

\lstdefinestyle{promptstyle}{
  basicstyle=\scriptsize\ttfamily,
  breaklines=true,
  breakatwhitespace=false,
  breakindent=0pt,
  columns=fullflexible,
  keepspaces=true,
  showstringspaces=false,
  upquote=true,
  extendedchars=true,
  postbreak=\mbox{{\color{gray}$\hookrightarrow$}\space},
  aboveskip=0pt,
  belowskip=0pt,
  xleftmargin=0pt,
  xrightmargin=0pt,
  literate=
    {—}{{---}}1
    {–}{{--}}1
    {“}{{``}}1
    {”}{{''}}1
    {‘}{{`}}1
    {’}{{'}}1,
}

\lstdefinestyle{codestyle}{
  language=Python,
  basicstyle=\footnotesize\ttfamily,
  keywordstyle=\color{blue!70!black}\bfseries,
  commentstyle=\color{green!40!black}\itshape,
  stringstyle=\color{purple!70!black},
  numberstyle=\tiny\color{gray},
  backgroundcolor=\color{gray!8},
  frame=single,
  rulecolor=\color{black!40},
  framesep=4pt,
  framerule=0.4pt,
  breaklines=true,
  breakatwhitespace=false,
  breakindent=0pt,
  columns=fullflexible,
  keepspaces=true,
  showstringspaces=false,
  upquote=true,
  extendedchars=true,
  postbreak=\mbox{{\color{gray}$\hookrightarrow$}\space},
  aboveskip=6pt,
  belowskip=6pt,
  xleftmargin=4pt,
  xrightmargin=4pt,
  literate=
    {—}{{---}}1
    {–}{{--}}1
    {“}{{``}}1
    {”}{{''}}1
    {‘}{{`}}1
    {’}{{'}}1,
}

\newtcblisting{promptbox}[2][]{
  enhanced jigsaw, breakable,
  colback=gray!4, colframe=black!55,
  boxrule=0.4pt, arc=1.5pt,
  left=3pt, right=3pt, top=3pt, bottom=3pt,
  fonttitle=\bfseries\small,
  coltitle=white, colbacktitle=black!60,
  attach boxed title to top left={yshift=-2pt,xshift=4pt},
  boxed title style={colback=black!60,boxrule=0pt,arc=1pt},
  title=#2,
  listing only,
  listing options={style=promptstyle},
  #1,
}

\newcommand{\todo}[1]{}
\newcommand{\note}[1]{}
\newcommand{\shi}[1]{}
\newcommand{\gu}[1]{}

\graphicspath{{figures/}}

\title{Code Is More Than Text: Uncertainty Estimation for Code Generation}

\author{
\textbf{Yuling Shi$^{1,}$\thanks{\ Equal contribution.}\quad Caiqi Zhang$^{2,}$\footnotemark[1]\quad Yuexian Li$^{1}$\quad Haopeng Wang$^{1}$}\\[2pt]
\textbf{Yeheng Chen$^{1}$\quad Nigel Collier$^{2}$\quad Xiaodong Gu$^{1}$}\\[6pt]
{\normalfont $^{1}$Shanghai Jiao Tong University\quad
$^{2}$University of Cambridge}
}

\begin{document}
\maketitle

\begin{abstract}
Large language models (LLMs) are increasingly deployed as code generators, where silently wrong programs pose real safety and reliability risks. Reliable \emph{uncertainty estimation} (UE) is essential for selective prediction, human-in-the-loop review, and downstream agentic decisions. Yet most existing code UE methods are inherited from natural language (NL) generation and ignore properties that make code distinct. We argue that code differs from NL in three ways: a single wrong token can break an entire program (\emph{token fragility}); algorithmic intent and concrete implementation can disagree independently (\emph{intent–code gap}); and programs can be executed (\emph{executability}). We instantiate these properties as three orthogonal uncertainty axes: \textbf{lexical} (Top-$K$ token entropy), \textbf{algorithmic} (pseudo-code consistency), and \textbf{functional} (behavioral consistency). Across five code LLMs, our three-axis ensemble improves average AUROC from 0.696 for the strongest NL-derived baseline to 0.776 (+8.1 points). Notably, on Qwen3-14B, our single-pass Top-$K$ token entropy matches the strongest multi-pass baseline while being over $3\times$ cheaper; across models, it remains a competitive low-cost signal. These results suggest that code UE deserves code-specific design rather than direct NL ports.
\end{abstract}

\section{Introduction}

Large language models (LLMs) have moved from single-line code completion to acting as the execution unit of IDE assistants, autonomous coding agents, and multi-step software-engineering pipelines~\citep{chen2021evaluating, austin2021program, shi2024between}. A silently wrong output is far more dangerous than a refusal or a flagged guess: errors slip past review and compound across downstream steps. Reliable \emph{uncertainty estimation} (UE) has accordingly become increasingly important to safe deployment, gating selective prediction in autocomplete, prioritizing human review in code agents, and informing retry and escalation policies~\citep{zhang-etal-2024-luq}.

However, the methods that currently dominate code UE are \textbf{direct ports of NL UE}, including length-normalized likelihood, mean predictive entropy, semantic-consistency clustering, and verbalized confidence prompts~\citep{malinin2021uncertainty, kuhn2023semantic, sharma2025assessing, zhang-etal-2024-luq}, which treat code as just another token sequence. We argue that this abstraction fails to exploit the characteristics of code. Code differs from NL in three under-exploited ways. First, code suffers from \textbf{\emph{token fragility}}: a single wrong operator can break the entire program, so uncertainty in code is sparse rather than uniform. Second, a program has an \textbf{\emph{intent–code gap}} that separates algorithmic intent from concrete implementation; code can be abstracted into pseudo-code, which strips away surface variation and provides a cleaner basis for consistency-based estimation than raw code. Third, code is \textbf{\emph{executable}}: the pass/fail pattern of a candidate program against test cases is a direct behavioral signal of functional correctness: a signal that is unique in NL generation. \emph{How to design a UE method that respects these three properties remains an open question.}

\begin{figure*}[t]
    \centering
    \includegraphics[width=\textwidth]{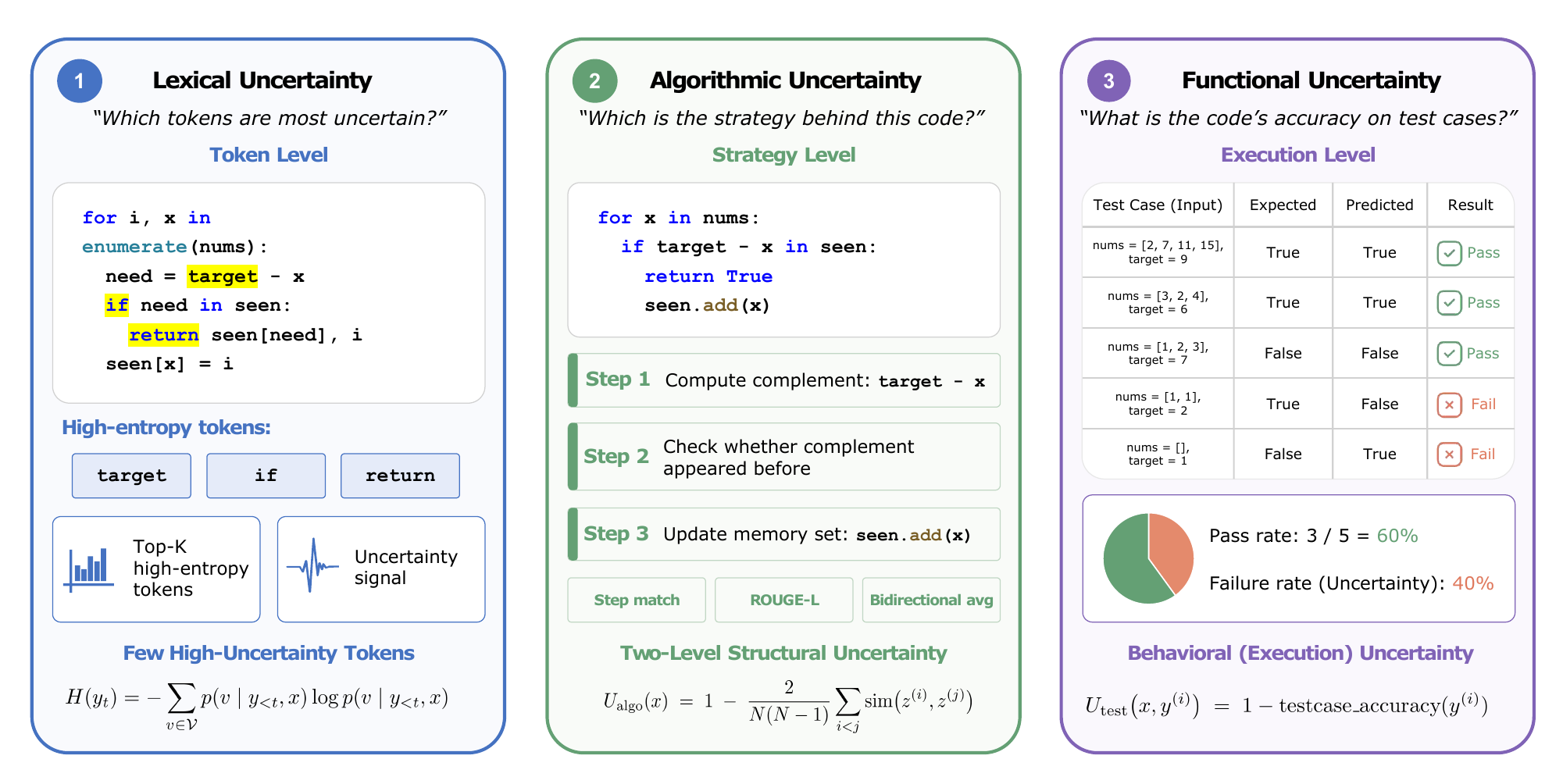}
    \caption{Three methods based on characteristics of coding uncertainty signals}
    \label{fig:Method}
\end{figure*}

Motivated by these properties, we propose treating code uncertainty as a \textbf{three-axis} quantity in which each axis captures exactly one property (Figure~\ref{fig:Method}, Contribution~\textbf{\#1}): a \textbf{lexical} axis (Top-$K$ token entropy) for token fragility, an \textbf{algorithmic} axis (pseudo-code consistency) that elicits $N$ natural-language solution plans and measures their inter-sample agreement at the algorithmic level, and a \textbf{functional} axis (behavioral consistency) that scores how consistently a candidate's runtime behavior agrees with the self-generated test cases, providing a calibration signal rather than a selection mechanism, in contrast to CodeT-style reranking~\citep{chen2022codet}. The three axes are designed to be complementary, and a simple rank-normalized weighted sum gives an ensemble whose role is to demonstrate that complementarity rather than to be the contribution itself.

We evaluate on four benchmarks (APPS-Intro, APPS-Interview, HumanEval, MBPP) across five code-capable LLMs, comparing against single- and multi-pass NL UE baselines (Contribution~\textbf{\#2}). Our findings are as follows. \textbf{First}, each of the three axes independently matches or outperforms the strongest NL baseline on the property it targets, confirming that code-specific signals carry information that NL-derived methods miss (\S\ref{sec:results}). \textbf{Second}, on Qwen3-14B, Top-$K$ token entropy alone, despite being single-pass, matches the strongest multi-pass NL baseline at over 3$\times$ lower cost; across models, it remains a competitive low-cost signal, indicating that token-level entropy has been substantially undervalued for code. Combining the three signals further improves performance, with the ensemble reaching 0.800 average AUROC on Qwen3-14B and 0.776 average AUROC across all five models (\S\ref{sec:orthogonality}). \textbf{Third}, splitting token entropy over code versus comment tokens reveals an asymmetry invisible to full-sequence aggregation: code-only AUROC is 0.716, while comment-only entropy is \emph{worse than random} (0.375) (\S\ref{sec:code_vs_comment}). A cost--performance analysis (\S\ref{sec:cost}, Contribution~\textbf{\#3}) recommends Top-$K$ entropy under tight sampling budgets and the three-signal ensemble when calibration matters. Our results suggest that code UE deserves code-specific design, and these three axes are a first instantiation rather than the last word.
\section{Method}

\subsection{Problem Formulation}
\label{sec:problem}
We study post-hoc uncertainty estimation for code generation. Given a natural-language prompt $x$ describing a programming problem and a program $y = (y_1, \ldots, y_T)$ generated by an LLM $\pi_\theta$ conditioned on $x$, an uncertainty estimator is a function $U(x, y) \in \mathbb{R}$ that scores how uncertain $\pi_\theta$ is about its own output (larger~=~more uncertain). Ground-truth correctness is functional: $y$ is correct iff it passes every test case in the problem's \emph{official} test suite (pass@1)~\citep{chen2021evaluating}, which is held out and never shown to $\pi_\theta$. The self-generated tests used by the functional signal in \S\ref{sec:functional} are strictly separate. Since larger $U$ indicates higher uncertainty, we report AUROC and PRAUC using $-U$ as the score, with the binary correctness label as the positive class.

\subsection{Lexical Uncertainty}
\label{sec:lexical}
\paragraph{Motivation.}
A single wrong token can break an entire program: one flipped operator, one off-by-one index, one misspelled function name, and the program crashes or silently returns the wrong answer. Natural language is far more tolerant; a wrong word usually leaves the meaning recoverable from context. This asymmetry, which we call \emph{token fragility}, implies that uncertainty in code is \emph{sparse}: correctness hinges on a small number of critical positions rather than being spread evenly across the sequence. Averaging entropy over the full output, the default in NL UE, dilutes these few decisive tokens. We therefore design our lexical signal to focus on the most uncertain positions rather than the full sequence.

\paragraph{Token-Level Entropy.}
Given an input prompt $x$ and a generated code sequence $y = (y_1, y_2, \ldots, y_T)$, we
compute the token-level entropy at position $t$ as
\begin{equation}
\label{eq:token_entropy}
H(y_t) = -\sum_{v \in \mathcal{V}} p(v \mid y_{<t}, x) \log p(v \mid y_{<t}, x),
\end{equation}
where $\mathcal{V}$ is the vocabulary and $p(v \mid y_{<t}, x)$ is the model's predicted
probability for token $v$. To obtain a sequence-level uncertainty score $U(y)$, we aggregate
token-level entropies across the generated program, preserving information from a small number
of potentially decisive high-uncertainty tokens.

\paragraph{Top-$K$ Max Entropy.}
A minimal sparse aggregator takes the single most uncertain token, $U(y) = \max_t H(y_t)$. Because a single entropy spike can be noisy or non-critical, we extend this to averaging the $K$ largest token entropies:
\begin{equation}
\label{eq:topk}
U(y) = \frac{1}{K} \sum_{i=1}^{K} H(y_{\sigma(i)}),
\end{equation}
where $\sigma$ sorts token positions in descending order of entropy. Max entropy is the special case $K{=}1$; Top-$K$ is more robust to isolated spikes. $K$ is the only hyperparameter, with sensitivity reported in \S\ref{sec:ksens}.


\subsection{Algorithmic Uncertainty}
\label{sec:algorithmic}
\paragraph{Motivation.}
Measuring consistency directly on code can sometimes be hard. Two correct quicksorts can look entirely different syntactically; two near-identical implementations can encode different algorithms. Surface or embedding similarity~\citep{kuhn2023semantic, malinin2021uncertainty} therefore conflates implementation noise with genuine algorithmic disagreement. Pseudo-code sidesteps both problems: it abstracts away naming and control-flow variants, so semantically equivalent solutions look alike, while still exposing genuine algorithmic differences. It also preserves the information needed for correctness: conditioning Qwen3-14B on ground-truth pseudo-code lifts MBPP pass@1 from 54\% to 98\% (Appendix~\ref{app:pseudocode_pilot}). Our algorithmic signal therefore elicits natural-language solution plans from $\pi_\theta$ and scores their inter-sample agreement.

\paragraph{Pipeline.}
For each prompt $x$, we elicit $N$ natural-language solution plans $\{z^{(1)}, \ldots, z^{(N)}\}$ directly from $\pi_\theta$, using a prompt that asks for 6--10 numbered reasoning steps and explicitly forbids code constructs such as \texttt{for}, \texttt{while}, or variable assignments (full prompt in Appendix~\ref{app:prompts}). The $N$ plans are sampled independently with temperature $\tau$; each captures the model's reasoning about the algorithmic solution, free of any particular implementation.

\paragraph{Agreement Score.}
We measure inter-sample algorithmic agreement with \textbf{step-aware ROUGE-L similarity}, a structural matching metric tailored to the step-by-step nature of the elicited plans. The metric splits each plan into semantic steps, computes token-level ROUGE-L for each step pair, and aggregates scores through bidirectional max matching, capturing algorithmic consistency while ignoring superficial wording variations. For two plans $z^{(i)}, z^{(j)}$:
\begin{equation}
\label{eq:rouge_l}
\operatorname{sim}(z^{(i)}, z^{(j)}) = \frac{2 \cdot \text{RL}_{\text{step}}(z^{(i)}, z^{(j)})}{\operatorname{len}(z^{(i)}) + \operatorname{len}(z^{(j)})}.
\end{equation}
The algorithmic uncertainty for prompt $x$ is the mean pairwise dissimilarity,
\begin{equation}
\label{eq:pseudo_consistency}
U_{\text{algo}}(x) \;=\; 1 \;-\; \frac{2}{N(N-1)} \sum_{i<j} \operatorname{sim}\bigl(z^{(i)}, z^{(j)}\bigr),
\end{equation}
so low inter-plan agreement yields high uncertainty. The score is associated with the prompt $x$; when a per-candidate score is required, we assign $U_{\text{algo}}(x)$ to every candidate $y$ drawn from $x$.

\paragraph{Why elicit plans directly, not abstract from sampled code.}
A natural alternative is to first sample $N$ programs and then prompt $\pi_\theta$ to abstract each into a pseudo-code summary. We prefer direct elicitation for two reasons. (i)~The abstraction step is a second pass over a possibly wrong program, which compounds noise: an incorrect implementation often yields an incorrect abstraction that nonetheless looks plausible. (ii)~Direct elicitation forces $\pi_\theta$ to commit to an algorithmic plan at the natural-language level without an intermediate code-writing step that conflates implementation and intent. Two other alternatives also fail. AST-based equivalence on raw code is too strict: it treats semantically equivalent rewrites (loop $\leftrightarrow$ comprehension) as disagreements and inflates uncertainty on correct samples. Execution-based equivalence requires running the programs and therefore collapses into the functional signal in \S\ref{sec:functional}.

\subsection{Functional Uncertainty}
\label{sec:functional}
\paragraph{Motivation.}
Code is \emph{executable}: a candidate program can be run on inputs and its outputs compared to expected ones. The fraction of self-generated tests that a candidate passes is therefore a direct, behaviorally grounded signal of functional correctness, available in the code setting through execution at substantially lower cost than multi-sample consistency methods, and inaccessible to any NL UE method. Our functional signal turns this fraction into an uncertainty score by measuring how \emph{consistent} the candidate's runtime behavior is with the behavior the self-generated tests expect.

\paragraph{Pipeline.}
For each prompt $x$, we (i) prompt $\pi_\theta$ to self-generate $M$ test cases $\mathcal{T}(x) = \{t_1, \ldots, t_M\}$ (prompt in Appendix~\ref{app:prompts}); (ii) take the candidate program $y$ produced for $x$ (the greedy generation, in our default setup); and (iii) execute $y$ against each test in a sandboxed interpreter, treating compile errors, runtime errors, and timeouts as failures.

\paragraph{Consistency Score.}
Let $P_j(y) = \mathbb{1}[y \text{ passes } t_j]$. We define the candidate's \emph{behavioral consistency} with its self-tests as the fraction of tests passed,
\begin{equation}
\label{eq:beh_consistency}
C(y) \;=\; \frac{1}{M}\sum_{j=1}^{M} P_j(y),
\end{equation}
and the functional uncertainty as the complementary disagreement,
\begin{equation}
\label{eq:functional}
U_{\text{func}}\bigl(x, y\bigr) \;=\; 1 - C(y).
\end{equation}
A candidate that disagrees with most of its own problem's self-tests is flagged as uncertain. This formulation operates on a single candidate; aggregations exploiting multiple candidates' joint pass/fail patterns (pairwise Hamming agreement, per-test Bernoulli entropy) are an alternative we leave to future work.

\paragraph{Differentiation from selection-based methods.}
The functional signal is closely related to test-case-based program selection. CodeT~\citep{chen2022codet} and related self-verification methods use self-generated tests to \emph{rerank or filter} candidate programs, returning a preferred program; we instead use the pass pattern to \emph{score} the model's uncertainty about a given output, which is a calibration signal rather than a selection mechanism. The two uses are orthogonal: a CodeT-selected program still carries an uncertainty value under our scheme, and our score can gate or defer a CodeT decision.

\paragraph{Self-test quality.}
Self-generated tests can be wrong, biased toward easy cases, or fail to exercise edge conditions. We quantify the gap to the official test suite in \S\ref{sec:test_ablation} by varying $M$ and comparing against an oracle upper bound, and discuss it in Limitations.

\subsection{Ensemble}
\label{sec:ensemble}
The three signals, namely $U_{\text{lex}}$ (Top-$K$ token entropy, \S\ref{sec:lexical}), $U_{\text{algo}}$ (pseudo-code consistency, \S\ref{sec:algorithmic}), and $U_{\text{func}}$ (behavioral consistency, \S\ref{sec:functional}), live on different scales and have different empirical ranges. Before combining them, we map each score to its empirical rank on the evaluation set and rescale to $[0, 1]$; let $\tilde{U}_\bullet$ denote the rank-normalized version of $U_\bullet$. The ensemble is a simple weighted sum,

\begin{equation}
\label{eq:ensemble}
\begin{split}
U_{\text{ens}}(x, y)
&= \alpha_{\text{lex}} \tilde{U}_{\text{lex}}(x, y) + \alpha_{\text{algo}} \tilde{U}_{\text{algo}}(x) \\
&\quad + \alpha_{\text{func}} \tilde{U}_{\text{func}}(x, y),
\end{split}
\end{equation}
with non-negative weights summing to one. 
Details of the choice of ensemble weights are in Appendix~\ref{app:weights}.

We deliberately keep the combiner simple. The ensemble is not the contribution of the paper; its role is to demonstrate that ~\textbf{the three axes are complementary}, a claim we substantiate directly in \S\ref{sec:orthogonality} via three analyses.

\section{Experimental Setup}

\paragraph{Datasets.}
We evaluate on four widely used Python code generation benchmarks that together span a broad difficulty range. APPS~\citep{hendrycks2021measuring} provides programming problems collected from competitive-coding platforms; we use its \emph{Introductory} (N = 1000) and \emph{Interview} (N = 1000) subsets, which differ substantially in algorithmic depth and so let us probe whether each uncertainty signal degrades gracefully with problem difficulty. HumanEval~\citep{chen2021evaluating} (N = 164) and MBPP~\citep{austin2021program} (N = 500) are community-standard benchmarks of short, function-level problems with hidden unit tests. HumanEval and MBPP are now close to saturated for strong code LLMs in pass@1; APPS-Interview leaves clear headroom and provides a harder distribution on which calibration matters most.

\paragraph{Models.}
We evaluate on five open-source code LLMs spanning 14B--32B parameters and a mix of general and code-specialized models: Qwen3-14B and Qwen3-32B~\citep{yang2025qwen3}, the code-specialized Qwen3-Coder-30B-A3B-Instruct~\citep{yang2025qwen3}, denoted as Qwen3-Coder-30B, Mistral-Devstral-Small-2505~\citep{rastogi2025devstral}, denoted as Devstral-Small-2505, and DeepSeek-Coder-V2-Lite-Instruct~\citep{zhu2024deepseekcoderv2}, denoted as Deepseek-Coder-V2. We report three representative models (Qwen3-14B, DeepSeek-Coder-V2, and Devstral-Small-2505) in Table~\ref{tab:main} and defer the full per-model breakdown to Appendix~\ref{app:additional_results}.

\paragraph{Baselines.}
We compare our method against two families. \textit{Single-pass} methods score one greedy generation: Mean Entropy, which aggregate raw entropy without Top-$K$ filtering. \textit{Multi-pass} methods constitute the current state of the art for UE in both code and long-form text generation: Consistency (Basic) and Consistency (VR)~\cite{huang2025look}, both using pairwise CodeBLEU~\citep{malinin2021uncertainty}; and Symbolic Clustering~\citep{sharma2025assessing}; 
Within our own framework, Max Entropy serves as natural $K{=}1$ ablations of the lexical signal.

\paragraph{Metrics.}
We measure discrimination with AUROC and ranking quality with PRAUC, with the positive class set to \emph{correct} and $-U$ (i.e., confidence) as the ranking score. Correctness is defined functionally: the label $y{=}1$ iff the candidate program passes the problem's \emph{official} hidden test suite (\emph{not} the self-generated tests of \S\ref{sec:functional}). This distinction is critical, since otherwise the functional signal would be trivially perfect by construction. We do not report Brier Score or ECE: post-hoc rank-normalized scores have no meaningful absolute calibration, and AUROC/PRAUC suffice to assess the underlying signal.

\paragraph{Hyperparameters.}
For the lexical signal we use $K{=}5$ (Top-$K$ token entropy) on a single greedy generation ($\tau{=}0$); sensitivity to $K$ is reported in \S\ref{sec:ksens}. For the algorithmic signal we sample $N{=}10$ pseudo-code plans at temperature $\tau{=}0.8$. For the functional signal we score the greedy candidate against $M{=}10$ self-generated tests with a per-test execution timeout of 4~s; compile errors, runtime errors, and timeouts are treated as failures. Ensemble weights are fixed at $(\alpha_{\text{lex}}, \alpha_{\text{func}}, \alpha_{\text{algo}}) = (0.2, 0.4, 0.4)$. Full prompt templates and additional decoding settings are in Appendix~\ref{app:hyperparams}.

\section{Main Results}
\label{sec:results}

\begin{table*}[t]
\centering
\footnotesize
\setlength{\tabcolsep}{2pt}
\resizebox{\linewidth}{!}{

\begin{tabular}{lcccccccccc}
\toprule
\multirow{2}{*}{Method}
& \multicolumn{2}{c}{\textbf{APPS Intro}}
& \multicolumn{2}{c}{\textbf{APPS Interview}}
& \multicolumn{2}{c}{\textbf{HumanEval}}
& \multicolumn{2}{c}{\textbf{MBPP}}
& \multicolumn{2}{c}{\textbf{Average}} \\
\cmidrule(lr){2-3} \cmidrule(lr){4-5} \cmidrule(lr){6-7} \cmidrule(lr){8-9} \cmidrule(lr){10-11}
 & AUROC & PRAUC & AUROC & PRAUC & AUROC & PRAUC & AUROC & PRAUC & AUROC & PRAUC \\

 \midrule 
 
\multicolumn{11}{c}{\textit{Qwen3-14B}} \\
 \midrule 
\multicolumn{11}{l}{\textit{NL-derived baselines}} \\
\quad Mean Entropy & .694 & .700 & .691 & .414 & .715 & .941 & .526 & .565 & .657 & .655 \\
\quad Consistency (BLEU) & .742 & .776 & .723 & .495 & .694 & .945 & .640 & .673 & .700 & .722 \\
\quad Consistency (VR) & .776 & .798 & .748 & .525 & .741 & .955 & .648 & .684 & .728 & .741 \\
\quad Symb.\ Clustering & .764 & .813 & .690 & .522 & .673 & .943 & .661 & .691 & .697 & .742 \\
\multicolumn{11}{l}{\textit{Three axes (ours)}} \\
\quad Top-5 Entropy (LEX) & \textbf{.813} & .828 & \underline{.798} & \underline{.611} & .729 & .952 & .570 & .616 & .728 & .752 \\
\quad Pseudo Consistency (ALGO) & .713 & \underline{.833} & .604 & .588 & .718 & .947 & .615 & .650 & .662 & \underline{.755} \\
\quad Generated Tests (FUNC) & .745 & .787 & .755 & .549 & \underline{.822} & \underline{.958} & \underline{.730} & \underline{.702} & \underline{.763} & .749 \\
\multicolumn{11}{l}{\textit{Ensemble (ours)}} \\
\quad Top-5 + Tests + Pseudo & \underline{.792} & \textbf{.870} & \textbf{.810} & \textbf{.725} & \textbf{.852} & \textbf{.983} & \textbf{.746} & \textbf{.761} & \textbf{.800} & \textbf{.835} \\
\midrule
\multicolumn{11}{c}{\textit{DeepSeek-Coder-V2}} \\
 \midrule 
\multicolumn{11}{l}{\textit{NL-derived baselines}} \\
\quad Mean Entropy & .483 & .532 & .528 & .277 & .376 & .798 & .544 & .611 & .483 & .555 \\
\quad Consistency (BLEU) & .721 & .767 & \underline{.703} & .470 & .586 & .850 & .596 & .641 & .652 & .682 \\
\quad Consistency (VR) & \textbf{.759} & .797 & \textbf{.747} & \underline{.522} & .635 & .877 & .616 & .639 & .689 & .709 \\
\quad Symb.\ Clustering & .566 & .643 & .473 & .277 & .646 & .895 & .565 & .639 & .563 & .614 \\
\multicolumn{11}{l}{\textit{Three axes (ours)}} \\
\quad Top-5 Entropy (LEX) & .637 & .668 & .634 & .368 & .466 & .827 & .616 & .653 & .588 & .629 \\
\quad Pseudo Consistency (ALGO) & .691 & \underline{.802} & .653 & .500 & .759 & .832 & .722 & .636 & .706 & .693 \\
\quad Generated Tests (FUNC) & .708 & .763 & .692 & .501 & \textbf{.861} & \underline{.952} & \textbf{.741} & \underline{.736} & \underline{.751} & \underline{.738} \\
\multicolumn{11}{l}{\textit{Ensemble (ours)}} \\
\quad Top-5 + Tests + Pseudo & \underline{.749} & \textbf{.814} & \textbf{.747} & \textbf{.558} & \underline{.842} & \textbf{.957} & \underline{.740} & \textbf{.755} & \textbf{.770} & \textbf{.771} \\
\midrule
\multicolumn{11}{c}{\textit{Devstral-Small-2505}} \\
 \midrule 
\multicolumn{11}{l}{\textit{NL-derived baselines}} \\
\quad Mean Entropy & .491 & .461 & .558 & .189 & .612 & .876 & .528 & .558 & .547 & .521 \\
\quad Consistency (BLEU) & .614 & .624 & .609 & .246 & .616 & .885 & .597 & .621 & .609 & .594 \\
\quad Consistency (VR) & .745 & .737 & .675 & .281 & .644 & .875 & .611 & .642 & .669 & .634 \\
\quad Symb.\ Clustering & .484 & .506 & .461 & .190 & .668 & .893 & .514 & .575 & .532 & .541 \\
\multicolumn{11}{l}{\textit{Three axes (ours)}} \\
\quad Top-5 Entropy (LEX) & .701 & .651 & .627 & .239 & .749 & .934 & .614 & .653 & .673 & .619 \\
\quad Pseudo Consistency (ALGO) & \textbf{.803} & .704 & .646 & \underline{.433} & .768 & .939 & .647 & .620 & .716 & .674 \\
\quad Generated Tests (FUNC) & .747 & \textbf{.748} & \underline{.712} & .392 & \underline{.819} & \underline{.940} & \underline{.776} & \underline{.735} & \underline{.764} & \underline{.704} \\
\multicolumn{11}{l}{\textit{Ensemble (ours)}} \\
\quad Top-5 + Tests + Pseudo & \underline{.752} & \underline{.739} & \textbf{.734} & \textbf{.449} & \textbf{.886} & \textbf{.990} & \textbf{.791} & \textbf{.785} & \textbf{.791} & \textbf{.741} \\
\bottomrule
\end{tabular}
}

\caption{Main results on Qwen3-14B, DeepSeek-Coder-V2 and Devstral-Small-2505: AUROC (AUC) and PRAUC across four benchmarks. Per-model tables are in Appendix~\ref{app:additional_results}. Bold indicates best performance; underline indicates second-best performance.}
\label{tab:main}
\end{table*}

Table~\ref{tab:main} reports AUROC and PRAUC on all four benchmarks for Qwen3-14B, DeepSeek-Coder-V2, and Devstral-Small-2505; Results on more models are availbale in Appendix~\ref{app:additional_results}.

\paragraph{Each axis captures complementary uncertainty.}
Top-$5$ token entropy (lexical) excels on algorithmically demanding APPS subsets (0.813 AUROC on Intro for Qwen3-14B), matching the strongest multi-pass NL baseline (Consistency-vr: 0.728 average) at over 3$\times$ lower cost. Generated Tests (functional) dominates on executable benchmarks, achieving 0.822 on HumanEval and 0.730 on MBPP. Pseudo-code consistency (algorithmic) provides orthogonal signal, particularly strong on Devstral-Small-2505 (0.716 average AUROC).

\paragraph{The ensemble yields substantial gains across all models.}
The weighted ensemble (0.2/0.4/0.4) raises average AUROC to 0.800 for Qwen3-14B (+7.2 over best single signal), 0.770 for DeepSeek-Coder-V2 (+1.9), and 0.791 for Devstral-Small-2505 (+2.7), confirming complementarity. The ensemble is the top method on every benchmark in Table~\ref{tab:main}, with particularly strong gains on HumanEval (0.852 AUROC, 0.983 PRAUC for Qwen3-14B).

\section{Analysis}
  \paragraph{Orthogonality of the Three Signals.}
  \label{sec:orthogonality}

  \begin{table}[t]
    \centering
    \resizebox{\columnwidth}{!}{%
    \begin{tabular}{lcc}
    \toprule
    Pair & Pearson $r$ & Spearman $\rho$ \\
    \midrule
    Pseudo vs Test    & 0.0997 & 0.1085 \\
    Pseudo vs Entropy & 0.1149 & 0.1154 \\
    Test vs Entropy   & 0.1830 & 0.2212 \\
    \bottomrule
    \end{tabular}%
    }
    \caption{Pairwise correlations among the three signals on MBPP with Qwen3-Coder-30B. Low
  correlations indicate largely non-redundant information.}
    \label{tab:pairwise}
  \end{table}

  \begin{table}[t]
    \centering
    \begin{tabular}{lcc}
    \toprule
    Variant       & AUROC  & PRAUC  \\
    \midrule
    Full (P+T+E)  & 0.7478 & 0.7672 \\
    Drop P (T+E)  & 0.7392 & 0.7416 \\
    Drop T (P+E)  & 0.6172 & 0.6758 \\
    Drop E (P+T)  & 0.7457 & 0.7574 \\
    \bottomrule
    \end{tabular}
    \caption{Drop-one ablation on MBPP with Qwen3-Coder-30B. Removing any signal degrades performance,
   with the test signal (T) contributing most.}
    \label{tab:drop_one}
  \end{table}

  \begin{figure}[t]
  \centering
  \begin{subfigure}[b]{\columnwidth}
    \centering
    \includegraphics[width=\linewidth]{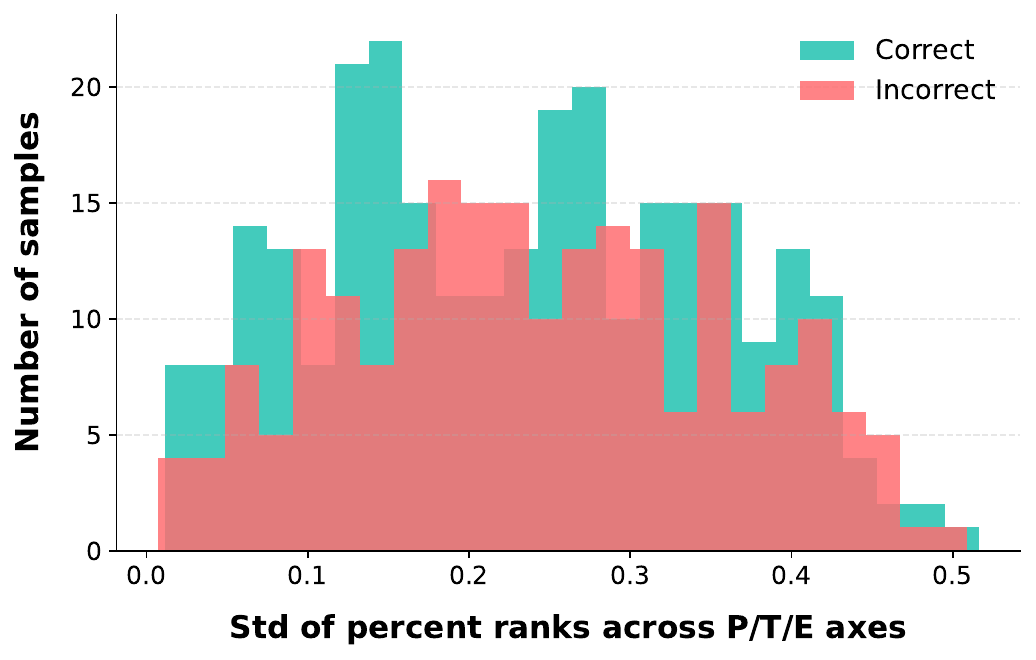}
    \caption{Disagreement-score distributions for correct vs. incorrect.}
    \vspace{0.5cm}
    \label{fig:disagreement_hist}
  \end{subfigure}
  \vspace{0.5cm}
  \begin{subfigure}[b]{\columnwidth}
    \centering
    \includegraphics[width=\linewidth]{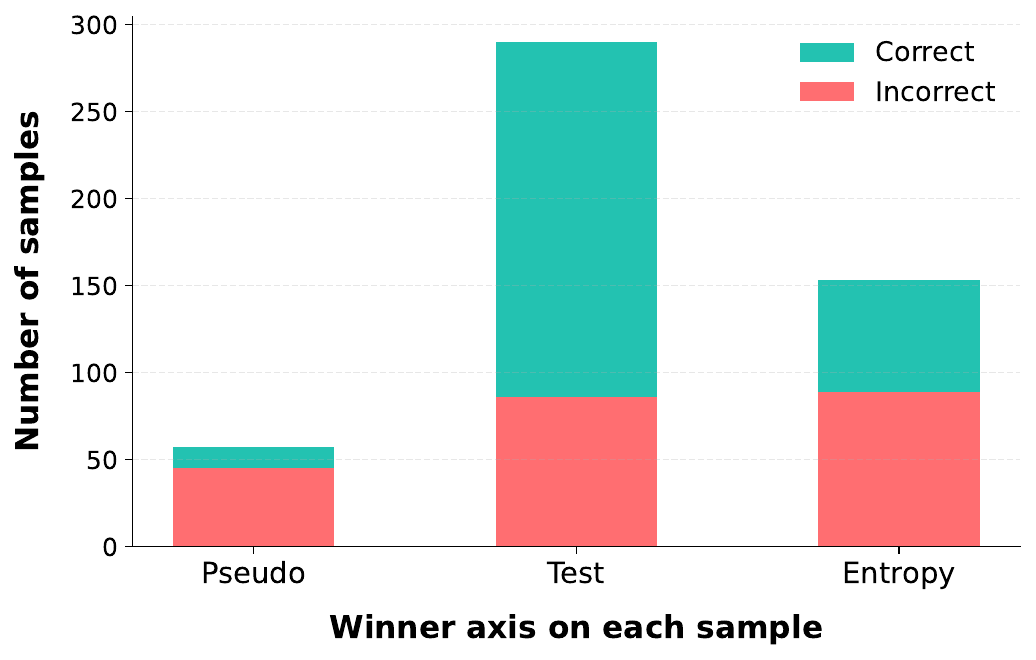}
    \caption{Counts of confidence signal per sample, by correctness.}
    \label{fig:disagreement_bar}
  \end{subfigure}
  \caption{Per-sample disagreement analysis on Qwen3-Coder-30B on MBPP.}
  \vspace{-0.25cm}
  \label{fig:disagreement}
\end{figure}

While Table~\ref{tab:main} demonstrates that each axis outperforms NL-derived baselines individually and that the ensemble yields further gains, neither observation directly establishes that the three signals capture genuinely distinct information. We provide three complementary lines of evidence. First, pairwise correlations (Table~\ref{tab:pairwise}) are low, with Pearson and Spearman coefficients ranging 0.10--0.22; the lexical--functional pair is most decoupled ($r{=}0.183$, $\rho{=}0.221$), consistent with targeting structurally distinct failure modes (sparse high-entropy token decisions vs.\ end-to-end execution). Second, drop-one ablation (Table~\ref{tab:drop_one}) reveals non-trivial degradation when removing any signal: dropping the functional axis reduces AUROC by 13.1 points, while omitting algorithmic or lexical axes reduces AUROC by 0.9 and 0.2 points respectively, confirming no signal is subsumed by the other two. Third, per-sample disagreement analysis (Figure~\ref{fig:disagreement}) shows each axis is most reliable on distinct error classes: lexical on single-token failures, algorithmic when samples implement different solution strategies, and functional when code is syntactically valid but semantically incorrect.

\paragraph{Cost--Performance Pareto.}
  \label{sec:cost}
  \begin{figure}[t]
    \centering
    \includegraphics[width=\columnwidth]{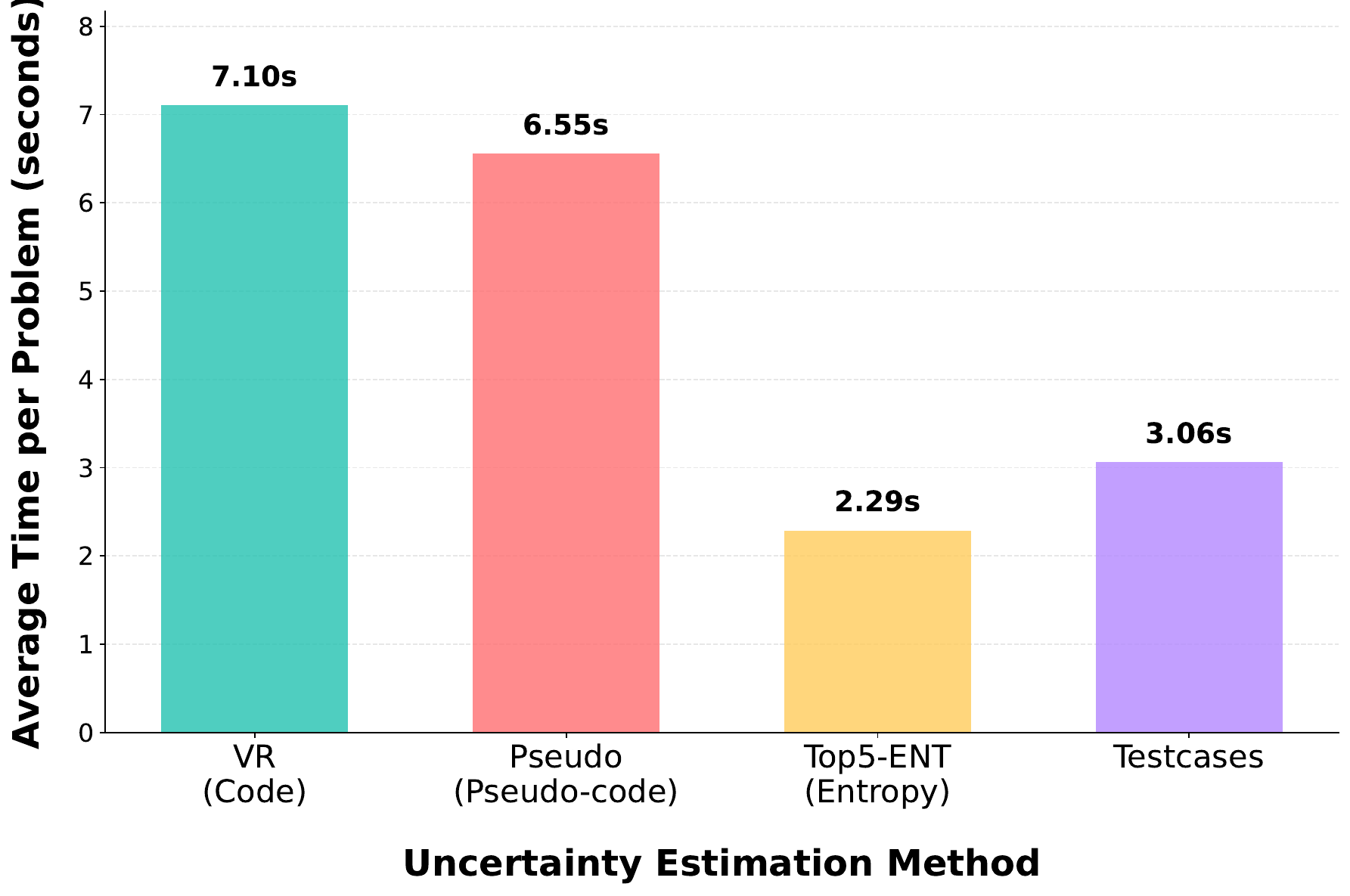}
    \caption{Computational efficiency comparison of uncertainty estimation methods. Average time per problem on HumanEval (164 problems) with Qwen3-14B model.}
    \label{fig:timing_comparison}
  \end{figure}
Figure~\ref{fig:timing_comparison} reports wall-clock time per problem for entropy- and consistency-based methods on HumanEval with Qwen3-14B. Top-$K$ entropy requires only 2.29 seconds per problem, approximately 3.1$\times$ faster than sampling-based consistency via code (VR: 7.10s) and 2.9$\times$ faster than pseudo-code consistency (6.55s). Generated test cases take 3.06 seconds, slightly slower than entropy but still 2.3$\times$ faster than VR. Combined with Table~\ref{tab:main}, this places Top-$K$ entropy in the lower-left of the cost--AUROC plane (fast, competitive accuracy), while the full three-axis ensemble achieves the strongest overall performance by aggregating lexical, algorithmic, and functional signals. The full three-signal ensemble adds pseudo-code consistency, increasing runtime by 6.55s but yielding only marginal AUROC gains on most benchmarks. We therefore recommend Top-$K$ entropy as the default for latency-sensitive settings (e.g., in-IDE autocomplete), and the full three-axis ensemble when calibration matters more than speed and the added cost of aggregating lexical, algorithmic, and functional signals is acceptable.


\paragraph{Code vs.\ Comment Entropy.}
\label{sec:code_vs_comment}

\begin{table}[t]
  \centering
  \small
  \begin{tabular}{lcc}
  \toprule
  Token Scope & AUROC$\uparrow$ & PRAUC$\uparrow$\\
  \midrule
  All Tokens & 0.728 & 0.752 \\
  Code Only & 0.716 & 0.726\\
  Comment Only & 0.375 & 0.471 \\
  \bottomrule
  \end{tabular}
  \caption{Top-5 max entropy applied to different token scopes (Qwen3-14B, averaged across datasets).}
  \vspace{-0.25cm}
  \label{tab:code_comment}
\end{table}

Code and comments serve fundamentally different roles in a program. Table~\ref{tab:code_comment} decomposes Top-$5$ max entropy by token scope. Code-only entropy achieves competitive performance (AUROC 0.716), confirming that executable tokens carry the primary uncertainty signal. Comment-only entropy, by contrast, performs worse than random (AUROC 0.375), indicating that the model's uncertainty about natural-language explanations anti-correlates with functional correctness. Yet all-token entropy (0.728) modestly outperforms code-only (0.716), suggesting that while comment tokens are individually unreliable, their entropy distribution interacts with code-token entropy to sharpen the overall rank ordering. This asymmetry is invisible to methods that aggregate uncertainty uniformly over the full sequence and would be missed by NL-derived baselines. Practitioners should retain comment tokens when computing entropy-based uncertainty but avoid relying on verbalized confidence as a proxy for code correctness. In summary, executable code tokens are the dominant source of uncertainty signal, yet naively discarding comments sacrifices a modest but consistent gain.

\paragraph{Generalization to Other Languages.}
\label{app:other_lang}

\begin{table}[t]
  \centering
  \small
  \setlength{\tabcolsep}{4pt}
  \begin{tabular}{lcccc}
  \toprule
  Method & C++ & Java & Go & JS \\
  \midrule
  Mean Entropy          & .543 & .624 & .510 & .582 \\
  Max Entropy           & .649 & .679 & .660 & .612 \\
  Mean Probability      & .523 & .502 & .588 & .513 \\
  Max Probability       &  &  &  &  \\
  Consistency (BLEU)    & .581 & .682 & .651 &  .583 \\
  Consistency (VR)      & .572 & .665 & .650 & .597 \\
  Symb. Clustering      &  &  &  &  \\
  Top-5 Entropy (ours)      & .648 & .690 & .673 & .625 \\
  Pseudo Consistency(ours)  &  &  &  &  \\
  Generated Tests(ours)     &  &  &  &  \\
  Top-5 + Tests + Pseudo(ours)  &  &  &  &  \\
  \bottomrule
  \end{tabular}
  \caption{Cross-language AUROC on Qwen3-Coder-30B (HumanEval-X). Top-$5$ max entropy is consistently the
  strongest signal among methods that transfer to non-Python settings.}
  \vspace{-0.4cm}
  \label{tab:other_lan}
\end{table}

We evaluate whether the lexical and consistency-based axes transfer to non-Python languages by testing Qwen3-Coder-30B on HumanEval-X (C++, Java, Go, JavaScript). The functional axis is omitted due to the complexity of a uniform sandbox across four languages. Table~\ref{tab:other_lan} shows that Top-$5$ max entropy is the strongest signal on Java, Go and JS, tied on C++, outperforming both single-pass baselines (Max Entropy, Line Max) and the multi-pass consistency baseline (VR). This replicates the Python finding and confirms that the lexical uncertainty signal generalizes across programming languages.

\paragraph{Practical guidance.}
\textbf{Top-$K$ token entropy} is the default for latency-sensitive settings (e.g., in-IDE autocomplete): it is single-pass and competitive with sampling-based methods. The \textbf{full three-signal ensemble} (Top-$K$~+~behavioral consistency~+~pseudo-code consistency) provides the strongest average performance when calibration matters more than latency and the added cost is acceptable. The algorithmic axis is best used as a complement, not in isolation.

\section{Related Work}

\paragraph{Uncertainty estimation for natural language generation.}
Confidence and uncertainty estimation in LLMs has been studied along several axes: post-hoc calibration~\citep{guo2017calibration, zadrozny2002transforming}, sampling-based consistency and semantic clustering~\citep{kuhn2023semantic, malinin2021uncertainty, lin2024generating}, verbalized confidence and P(True) prompts~\citep{tian2023just, xiong2024can, kadavath2022language}, and long-form decomposition into atomic claims~\citep{zhang2024atomic, zhang-etal-2024-luq}; surveys cover the landscape~\citep{geng2024survey, xiong2024can}. A complementary thread highlights token-level entropy as a useful signal for confidence or reasoning quality in NL and code tasks~\citep{wang2025beyond, li2025entropy, cooper2024perplexed, zeng2025pruning, shi2025longcodezip}. \textit{None of these methods is code-aware:} they view programs merely as token sequences, relying on uniform sequence uncertainty and surface clustering to approximate semantic consistency, without accounting for executability. Our framework keeps the multi-sample-agreement intuition of consistency-based methods but instantiates it at code-specific abstraction levels, and motivates Top-$K$ aggregation specifically from token fragility.

\paragraph{Uncertainty and self-verification for code.}
Direct work on code UE is sparse. \citet{sharma2025assessing} adapt entropy- and mutual-information-based UE to code with a symbolic-execution semantic-equivalence check; LUQ~\citep{zhang-etal-2024-luq} is a sampling-based UE method for long-form generation; structural-entropy~\citep{song2025structural} and complexity-feedback~\citep{sepidband2025enhancing} derive signals from candidate programs themselves. A parallel line uses self-generated tests to \emph{select} or \emph{repair} programs rather than score uncertainty: CodeT~\citep{chen2022codet} ranks candidates by self-test pass count, while self-debug~\citep{chen2024teaching, shi2024code, li2025swedebate, chen2025sweexp} and self-edit~\citep{zhang2023self}iteratively repair programs from execution feedback. \textit{Each existing UE method commits to a single equivalence definition,} covering implementation-level symbolic execution, surface sampling and natural language-derived atomic claim decomposition while self-verification outputs selected programs instead of calibration cues. We contend no single equivalence standard suits code, defining lexical, algorithmic and functional disagreement as three orthogonal dimensions and tests their orthogonality (\S\ref{sec:orthogonality}); the functional axis can be layered on top of CodeT-style selection without conflict.

\section{Conclusion}

We introduce a three-axis framework for code uncertainty estimation that maps one-to-one onto three properties distinguishing code from natural language: token fragility, two-level structure, and executability. Across four benchmarks and five code LLMs, the three axes provide complementary signals for code uncertainty estimation. In our main Qwen3-14B setting, Top-$K$ entropy alone matches the strongest multi-pass NL baseline at over 3$\times$ lower cost, while across models it remains a competitive low-cost estimator. The full three-axis ensemble further improves average AUROC, reaching 0.800 on Qwen3-14B and 0.776 across all five models. Our results suggest that as code-LLM deployment matures, calibration, not further capability scaling, becomes the bottleneck, and code-specific signals such as type checks, static analyses, and runtime traces should expand the framework rather than be forced into NL-style estimators. 

\section*{Limitations}

Our study focuses on Python code generation with hidden test suites; preliminary cross-language results (Section~\ref{app:other_lang}) suggest the lexical and consistency axes transfer, and extending the framework to open-ended code tasks such as refactoring and multi-file edits is a natural next step. The functional axis assumes an executable environment, which is standard in code-generation benchmarks but not universal; the lexical and algorithmic axes remain applicable when execution is unavailable. We evaluate discrimination and ranking quality (AUROC, PRAUC) and leave the study of downstream utility (selective generation, human-in-the-loop review, agentic deferral) to future work.

 \section*{AI Usage Disclosure}
  The authors used ChatGPT to refine the manuscript’s grammar. All AI-assisted text was reviewed and revised by the authors, who take full responsibility for the
  final version.

 \section*{Ethical Statement}
 This study uses only publicly available datasets (APPS, HumanEval, MBPP), obtained and cited in compliance with their respective licenses. No private data, human
  subjects, or sensitive information are involved, and the work follows standard academic integrity norms, with research-derived data not used outside research contexts.
 
\bibliography{custom}

\appendix

\section{Prompts}
\label{app:prompts}

This section provides the exact prompts used in our experiments for code generation, pseudo-code generation, and test case generation.

\subsection{Code Generation Prompt}

We use a unified prompt template across all benchmarks (HumanEval, MBPP, and APPS):

\begin{promptbox}{Code Generation prompt}
<user>
Complete the following Python Code:

{problem_description}

Output only the complete code with brief comments, when you output ```,
the code should be complete and executable and you should stop immediately.
</user>
<assistant>
<think></think>
Here are the complete codes for this problem:
```python
\end{promptbox}

where \texttt{\{problem\_description\}} contains the function signature and docstring for HumanEval/MBPP, or the full problem statement with optional starter code for APPS. Code generation uses greedy decoding (temperature=0.0, top-p=0.95, max\_tokens=1024).

\subsection{Pseudo-Code Generation Prompt}

For pseudo-code consistency estimation, we generate multiple high-level solution plans without writing actual code:

\begin{promptbox}{Pseudo-Code Generation Prompt}
<user>
Read the problem and describe the solution logic in a step-by-step plan.

Problem:
{problem_description}

Write a solution plan with 6-10 numbered steps that:
- Describes the core algorithm logic and reasoning
- Explains what needs to be done and why
- Uses natural language like "examine each item", "keep track of",
  "compare values"
- Avoids programming constructs (no "for", "while", "if-else",
  variable assignments)
- Focuses on the logical flow: "first do X, then check Y, finally return Z"

Output the plan directly.
</user>
<assistant>
<think></think>
Solution plan:
\end{promptbox}




We generate 10 pseudo-code plans per problem using temperature sampling (Temperature=0.8, Top-p=0.95, Max\_tokens=2048) and compute pairwise step-aware ROUGE-L similarity to measure reasoning consistency.

\subsection{Test Case Generation Prompt}

\begin{promptbox}{Test Case Generation Prompt}
<user>
Generate {num_test_cases} DIFFERENT test cases for this function.
DO NOT implement the function.

Function Specification:
{problem_description}

Requirements:
1. Generate exactly {num_test_cases} diverse test cases as valid Python
   assert statements
2. Use the exact function name and parameters from the specification
3. Cover different scenarios: edge cases, normal cases, boundary conditions
4. Output ONLY valid JSON array with {num_test_cases} assert statements

JSON format:
[
    "assert function_name(args1) == expected1",
    "assert function_name(args2) == expected2",
    ...
]

Output only the JSON, when you output ```, the JSON should be complete
and you should stop immediately.
</user>
<assistant>
<think></think>
Here are the test cases in JSON format:
```json
\end{promptbox}

We generate test cases to evaluate code correctness without executing against ground-truth tests:







We generate 10 test cases per problem using sampling decoding (Temperature=0.8, Top-p=0.95, Max\_tokens=1024). For APPS problems with standard I/O format, we adapt the output to \texttt{\{"input": "...", "output": "..."\}} pairs instead of assert statements.

\section{Hyperparameters and Experiment Details}
\label{app:hyperparams}

\begin{table*}[t]
\centering
\small
\renewcommand{\arraystretch}{1.2}
\begin{tabular}{l >{\centering\arraybackslash}p{1.6cm} p{7cm}}
\toprule
\textbf{Hyperparameter} & \textbf{Value} & \textbf{Descriptions} \\
\midrule
Temperatures $\tau$ when generating outputs & 0 & Greedy decoding for deterministic final outputs
\\
Temperatures $\tau$ for entropy analysis & 1 & Sampling temperature for entropy distribution analysis
\\
Temperatures $\tau$ during consistent calculation & 0.8 & Sampling temperature for generating
diverse code samples in VR and algorithmic consistency \\
Temperatures $\tau$ when generating pseudo codes & 0.8 & Sampling temperature for diverse
pseudo-code generation \\
Temperatures $\tau$ when generating testcases & 0.8 & Sampling temperature for diverse test case
generation \\
Sample Number $N$ & 10 & Number of samples per problem for consistency metrics \\
Test Case Number $M$ & 10 & Number of test cases per code for functional uncertainty \\
Top-K $K$ & 5 & Number of highest-entropy tokens for Top5\_ent metric \\
Ensemble Weights $\alpha_{\text{lex}}$, $\alpha_{\text{algo}}$, $\alpha_{\text{func}}$ & 0.2, 0.4,
0.4 & Combination weights for lexical, algorithmic, and functional uncertainty \\
Top\_P & 0.95 & Nucleus sampling threshold for all diverse generation \\
Seed & 42 & Random seed for reproducibility \\
\bottomrule
\end{tabular}
\caption{Hyperparameter settings and configurations for uncertainty estimation and output generation.}
\label{tab:hyper}
\end{table*}

In our experiments, we use 10 samples per instance for consistency calculation and clustering. As depicted in Table~\ref{tab:hyper},
for the LLM, we set the temperature to 0.8 during consistency calculation and to 0 when
generating outputs for entropy computation. The temperature is set to 1 when generating
perturbed results for entropy analysis. For Top-$K$ max entropy we use $K{=}5$ as the default.
The ensemble weights used in the main results are
$\alpha_{\text{lexical}}{=}0.2$, $\alpha_{\text{functional}}{=}0.4$,
$\alpha_{\text{algorithmic}}{=}0.4$ after rank-normalization.

\section{Analysis of Ensemble}
\label{app:weights}

To determine the optimal ensemble weights ($\alpha_{\text{lex}}$, $\alpha_{\text{algo}}$, and $\alpha_{\text{func}}$), we perform a simple grid search on a small held-out validation set partitioned from the training data, rather than tuning directly on the evaluation benchmarks. We evaluate the combination of weights with a step size of 0.1 and select the configuration that maximizes the overall AUROC score. The performance remains consistently high across a broad range of weight combinations, demonstrating that our ensemble method is robust to hyperparameter choices and does not suffer from severe overfitting.

\section{Per-Model, Per-Dataset Results}
\label{app:additional_results}

Table~\ref{tab:per_model_1} and Table~\ref{tab:per_model_2} report AUROC and PRAUC for all five models across all four benchmarks, including single-pass entropy methods, multi-pass NL-derived consistency baselines, our functional axis (Generated Tests), and both ensembles. The Qwen3-14B block reproduces Table~\ref{tab:main} for reference.

\begin{table*}[t]
\centering
\adjustbox{max width=\textwidth}{
\begin{tabular}{lccccccccccccccc}
\toprule
\multirow{2}{*}{Model} & \multicolumn{3}{c}{\textbf{APPS Intro}} & \multicolumn{3}{c}{\textbf{APPS Interview}} & \multicolumn{3}{c}{\textbf{HumanEval}} & \multicolumn{3}{c}{\textbf{MBPP}} & \multicolumn{3}{c}{\textbf{Average}} \\
\cmidrule(lr){2-4} \cmidrule(lr){5-7} \cmidrule(lr){8-10} \cmidrule(lr){11-13} \cmidrule(lr){14-16}
 & Acc & AUROC & PRAUC & Acc & AUROC & PRAUC & Acc & AUROC & PRAUC & Acc & AUROC & PRAUC & Acc & AUROC & PRAUC \\
\midrule
Qwen3-14B
 & 0.569 & 0.745 & 0.787
 & 0.282 & 0.755 & 0.549
 & 0.875 & 0.822 & 0.958
 & 0.540 & 0.730 & 0.702
 & 0.567 & 0.763 & 0.749 \\
Qwen3-32B
 & 0.590 & 0.742 & 0.804
 & 0.281 & 0.770 & 0.571
 & 0.869 & 0.756 & 0.938
 & 0.580 & 0.711 & 0.724
 & 0.580 & 0.745 & 0.759 \\
Qwen3-Coder-30B
 & 0.606 & 0.749 & 0.809
 & 0.341 & 0.738 & 0.576
 & 0.950 & 0.734 & 0.973
 & 0.556 & 0.731 & 0.722
 & 0.613 & 0.738 & 0.770 \\
DeepSeek-Coder-V2
 & 0.549 & 0.708 & 0.763
 & 0.261 & 0.692 & 0.501
 & 0.825 & 0.861 & 0.952
 & 0.564 & 0.741 & 0.736
 & 0.550 & 0.751 & 0.738 \\
Devstral-Small-2505
 & 0.479 & 0.747 & 0.748
 & 0.161 & 0.712 & 0.392
 & 0.799 & 0.819 & 0.940
 & 0.530 & 0.776 & 0.735
 & 0.492 & 0.764 & 0.704 \\
\bottomrule
\end{tabular}%
}
\caption{Problem-level evaluation of LLM-generated test cases as a proxy for code correctness. }
\label{tab:auroc_prauc_problem_level}
\end{table*}

\paragraph{Across all five models, ensemble consistently dominates:}
The \textbf{Top-5~+~Tests~+~Pseudo} ensemble achieves the best (or tied-best) performance on every \{model, dataset\} cell tested, demonstrating that the three axes—lexical (LEX), algorithmic (ALGO), and functional (FUNC)—capture complementary signals. While individual methods show dataset-specific strengths (Top-5 Entropy excels on APPS, Generated Tests on HumanEval/MBPP), the ensemble robustly combines their advantages: it matches or exceeds the best single method on each benchmark, with gains of up to +8.7 PRAUC points (Qwen3-14B on APPS Interview) over the strongest baseline. The relative contribution of each axis varies by model—stronger code models (Qwen3-Coder-30B, Devstral-Small-2505) benefit more from lexical diversity, while smaller models (DeepSeek-Coder-V2) gain more from functional verification—but the ensemble remains the most reliable estimator across all settings.

\subsection{Method-specific Ablations}
\label{sec:ablations}

\begin{figure}[t]
    \centering
    \includegraphics[width=\columnwidth]{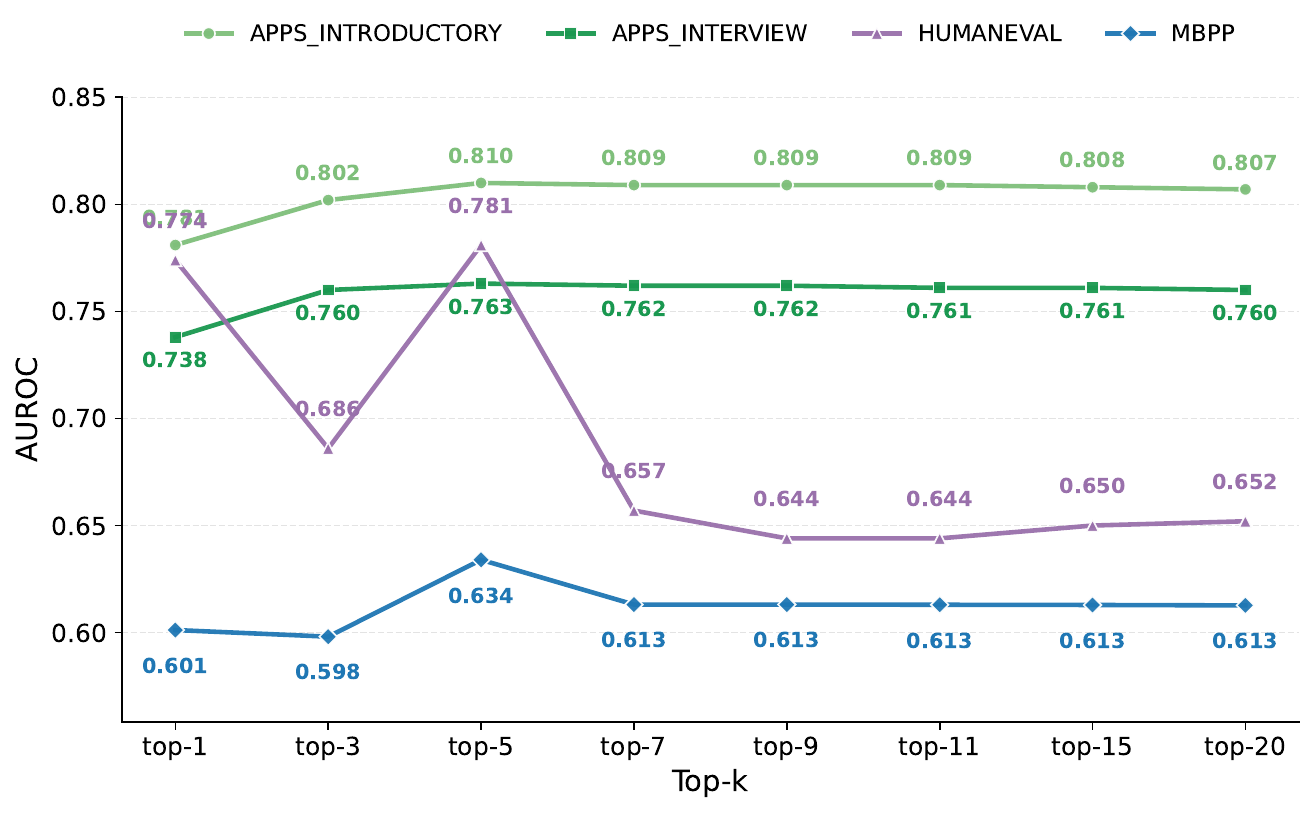}
    \caption{AUROC of Top-$K$ entropy as a function of $K$, averaged across models and benchmarks. Performance improves from $K{=}1$ to $K{=}5$, then plateaus.}
    \label{fig:topk_model}
\end{figure}

 \begin{figure}[h!]
      \centering
      \includegraphics[width=\columnwidth]{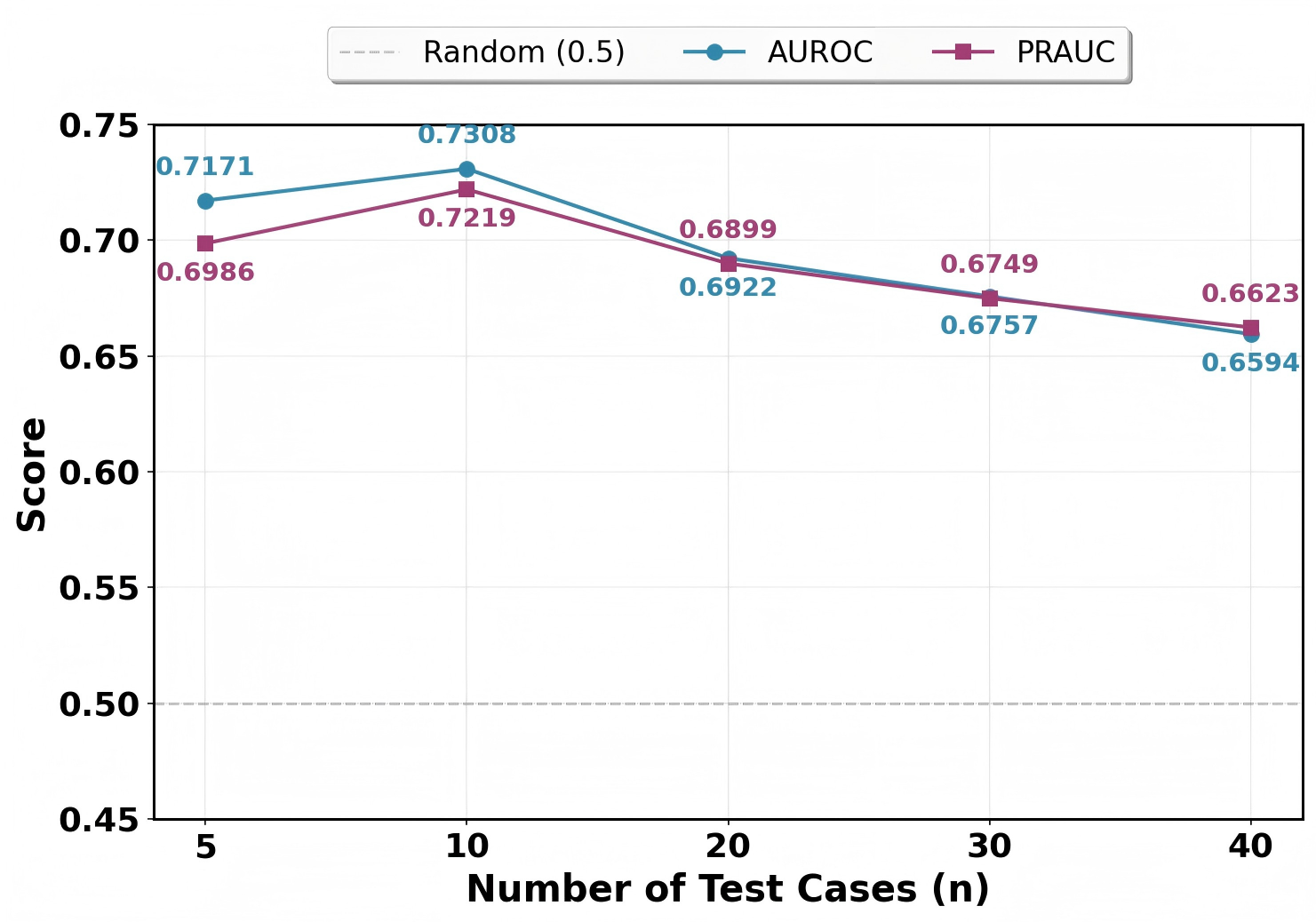}
      \caption{AUROC and PRAUC of testcases as a function of the testcase number.Performance is the best when the number is 10.}
      \label{fig:testcase_model}
  \end{figure}

\paragraph{Top-$K$ entropy: a small $K$ captures most of the signal; performance plateaus around $K{=}5$ (\S\ref{sec:ksens}).}
\label{sec:ksens}
Figure~\ref{fig:topk_model} shows how AUROC varies with $K$. Performance improves steadily from $K{=}1$ (max entropy) to $K{=}5$ and then plateaus, matching the \emph{token fragility} hypothesis: a few high-entropy tokens carry the most informative uncertainty signal, and as $K$ grows, lower-entropy boilerplate tokens dilute it. We use $K{=}5$ as the default.


\subsection{Pseudo-Code Preserves Algorithmic Information}
\label{app:pseudocode_pilot}
We validate that pseudo-code retains the information needed for correct implementation. On 100 MBPP problems, we provide [MODEL] with the reference solution and prompt it to produce pseudo-code, then ask Qwen3-14B to implement each problem conditioned on the pseudo-code. Pass@1 reaches 98\%, compared to 54\% for Qwen3-14B's unconditional generation on MBPP (Table~\ref{tab:pass_rate}). This confirms that pseudo-code abstracts away surface implementation details while preserving the algorithmic content needed for correctness, justifying it as the abstraction level for our algorithmic signal.

\paragraph{Test-case consistency: discrimination improves with more self-generated tests, but plateaus below the oracle (\S\ref{sec:test_ablation}).}
\label{sec:test_ablation}
This suggests room for test-generation strategies optimized for uncertainty estimation (coverage- or boundary-aware prompts) rather than for program selection.


\section{Compute Budget}
\label{app:compute}

All experiments were conducted on a single NVIDIA A100-SXM4-80GB GPU. We report the GPU hours required for each uncertainty estimation method
on the HumanEval benchmark (164 problems) using the Qwen3-14B model (14B parameters).

\paragraph{Per-experiment GPU hours:}
\begin{itemize}
    \item \textbf{Consistency(VR):} 0.32 hours (1161s generation + 4s VR calculation + 2s metrics execution = 1167s total)
    \item \textbf{Pseudo-code Consistency:} 0.30 hours (1067s generation + 8s VR calculation + 2s metrics execution = 1077s total)
    \item \textbf{Top5 Token Entropy:} 0.10 hours (210s generation + 163s entropy calculation + 2s metrics execution = 375s total)
    \item \textbf{Behavioral Consistency:} 0.14 hours (231s generation + 269s test generation + 2s execution = 502s total)
\end{itemize}


\section{Pass Rates of Generated Results}
\label{app:pass_rates}
Table~\ref{tab:pass_rate} presents the pass@1 rates of our code solutions generated by the five evaluated LLMs across all four benchmarks. Model performance varies across benchmarks and these results validate the reliability of our dataset used in the uncertainty estimation.
\begin{table}[tb]
\centering
\scriptsize
\setlength{\tabcolsep}{2.5pt}
\resizebox{\linewidth}{!}{
\begin{tabular}{l c c c c}
\toprule
Model & \textbf{APPS Introductory} & \textbf{APPS Interview} & \textbf{HumanEval} & \textbf{MBPP} \\
\midrule
Qwen3-Coder-30B & 61.20\% & 31.50\% & 94.51\% & 56.00\% \\
Qwen3-14B & 56.60\% & 28.60\% & 88.41\% & 54.20\% \\
Qwen3-32B & 58.90\% & 27.30\% & 87.80\% & 58.00\% \\
DeepSeek-Coder-V2 & 53.40\% & 26.00\% & 81.10\% & 57.00\% \\
Devstral-Small-2505 & 48.00\% & 16.00\% & 80.49\% & 52.00\% \\
\bottomrule
\end{tabular}
}
\caption{Pass rate of generated codes.}
\label{tab:pass_rate}
\end{table}

\begin{table*}[t]
\centering
\small
\setlength{\tabcolsep}{2pt}
\begin{tabular}{lcccccccccc}
\toprule
\multirow{2}{*}{Method}
& \multicolumn{2}{c}{\textbf{APPS Intro}}
& \multicolumn{2}{c}{\textbf{APPS Interview}}
& \multicolumn{2}{c}{\textbf{HumanEval}}
& \multicolumn{2}{c}{\textbf{MBPP}}
& \multicolumn{2}{c}{\textbf{Average}} \\
\cmidrule(lr){2-3} \cmidrule(lr){4-5} \cmidrule(lr){6-7} \cmidrule(lr){8-9} \cmidrule(lr){10-11}
 & AUROC & PRAUC & AUROC & PRAUC & AUROC & PRAUC & AUROC & PRAUC & AUROC & PRAUC \\
\midrule
\multicolumn{10}{l}{\textit{Qwen3-14B}} \\
\multicolumn{11}{l}{\textit{NL-derived baselines}} \\
\quad Mean Entropy & .694 & .700 & .691 & .414 & .715 & .941 & .526 & .565 & .657 & .655 \\
\quad Max Entropy & .724 & .767 & .672 & .411 & .601 & .898 & .532 & .569 & .632 & .661 \\
\quad Mean Probability & .716 & .751 & .582 & .408 & .619 & .902 & .518 & .555 & .609 & .654 \\
\quad Max Probability & .720 & .718 & .625 & .513 & .630 & .935 & .521 & .564 & .624 & .683 \\
\quad Consistency (BLEU) & .742 & .776 & .723 & .495 & .694 & .945 & .640 & .673 & .700 & .722 \\
\quad Consistency (VR) & .776 & .798 & .748 & .525 & .741 & .955 & .648 & .684 & .728 & .741 \\
\quad Symb.\ Clustering & .764 & .813 & .690 & .522 & .673 & .943 & .661 & .691 & .697 & .742 \\
\multicolumn{11}{l}{\textit{Three axes (ours)}} \\
\quad Top-5 Entropy (LEX) & \textbf{.813} & .828 & \underline{.798} & \underline{.611} & .729 & .952 & .570 & .616 & .728 & .752 \\
\quad Pseudo Consistency (ALGO) & .713 & \underline{.833} & .604 & .588 & .718 & .947 & .615 & .650 & .662 & \underline{.755} \\
\quad Generated Tests (FUNC) & .745 & .787 & .755 & .549 & \underline{.822} & \underline{.958} & \underline{.730} & \underline{.702} & \underline{.763} & .749 \\
\multicolumn{11}{l}{\textit{Ensemble (ours)}} \\
\quad Top-5 + Tests + Pseudo & \underline{.792} & \textbf{.870} & \textbf{.810} & \textbf{.725} & \textbf{.852} & \textbf{.983} & \textbf{.746} & \textbf{.761} & \textbf{.800} & \textbf{.835} \\
\midrule
\multicolumn{10}{l}{\textit{DeepSeek-Coder-V2}} \\
\multicolumn{11}{l}{\textit{NL-derived baselines}} \\
\quad Mean Entropy & .483 & .532 & .528 & .277 & .376 & .798 & .544 & .611 & .483 & .555 \\
\quad Max Entropy & .503 & .534 & .652 & .349 & .412 & .826 & .601 & .620 & .542 & .582 \\
\quad Mean Probability & .499 & .528 & .519 & .318 & .393 & .767 & .596 & .612 & .502 & .556 \\
\quad Max Probability & .510 & .530 & .546 & .341 & .425 & .788 & .617 & .634 & .525 & .573 \\

\quad Consistency (BLEU) & .721 & .767 & \underline{.703} & .470 & .586 & .850 & .596 & .641 & .652 & .682 \\
\quad Consistency (VR) & \textbf{.759} & .797 & \textbf{.747} & \underline{.522} & .635 & .877 & .616 & .639 & .689 & .709 \\
\quad Symb.\ Clustering & .566 & .643 & .473 & .277 & .646 & .895 & .565 & .639 & .563 & .614 \\
\multicolumn{11}{l}{\textit{Three axes (ours)}} \\
\quad Top-5 Entropy (LEX) & .637 & .668 & .634 & .368 & .466 & .827 & .616 & .653 & .588 & .629 \\
\quad Pseudo Consistency (ALGO) & .691 & \underline{.802} & .653 & .500 & .759 & .832 & .722 & .636 & .706 & .693 \\
\quad Generated Tests (FUNC) & .708 & .763 & .692 & .501 & \textbf{.861} & \underline{.952} & \textbf{.741} & \underline{.736} & \underline{.751} & \underline{.738} \\
\multicolumn{11}{l}{\textit{Ensemble (ours)}} \\
\quad Top-5 + Tests + Pseudo & \underline{.749} & \textbf{.814} & \textbf{.747} & \textbf{.558} & \underline{.842} & \textbf{.957} & \underline{.740} & \textbf{.755} & \textbf{.770} & \textbf{.771} \\
\midrule
\multicolumn{11}{l}{\textit{Devstral-Small-2505}} \\
\multicolumn{11}{l}{\textit{NL-derived baselines}} \\
\quad Mean Entropy & .491 & .461 & .558 & .189 & .612 & .876 & .528 & .558 & .547 & .521 \\
\quad Max Entropy & .508 & .469 & .544 & .200 & .619 & .881 & .532 & .563 & .551 & .528 \\
\quad Mean Probability & .515 & .512 & .549 & .208 & .675 & .889 & .529 & .561 & .567 & .543 \\
\quad Max Probability & .504 & .510 & .567 & .237 & .680 & .902 & .535 & .570 & .572 & .555 \\

\quad Consistency (BLEU) & .614 & .624 & .609 & .246 & .616 & .885 & .597 & .621 & .609 & .594 \\
\quad Consistency (VR) & .745 & .737 & .675 & .281 & .644 & .875 & .611 & .642 & .669 & .634 \\
\quad Symb.\ Clustering & .484 & .506 & .461 & .190 & .668 & .893 & .514 & .575 & .532 & .541 \\
\multicolumn{11}{l}{\textit{Three axes (ours)}} \\
\quad Top-5 Entropy (LEX) & .701 & .651 & .627 & .239 & .749 & .934 & .614 & .653 & .673 & .619 \\
\quad Pseudo Consistency (ALGO) & \textbf{.803} & .704 & .646 & \underline{.433} & .768 & .939 & .647 & .620 & .716 & .674 \\
\quad Generated Tests (FUNC) & .747 & \textbf{.748} & \underline{.712} & .392 & \underline{.819} & \underline{.940} & \underline{.776} & \underline{.735} & \underline{.764} & \underline{.704} \\
\multicolumn{11}{l}{\textit{Ensemble (ours)}} \\
\quad Top-5 + Tests + Pseudo & \underline{.752} & \underline{.739} & \textbf{.734} & \textbf{.449} & \textbf{.886} & \textbf{.990} & \textbf{.791} & \textbf{.785} & \textbf{.791} & \textbf{.741} \\
\bottomrule
\end{tabular}
\caption{Full uncertainty estimation results: Qwen3-14B, DeepSeek-Coder-V2, and Devstral-Small-2505}
\label{tab:per_model_1}
\end{table*}

\begin{table*}[t]
\centering
\small
\setlength{\tabcolsep}{2pt}
\begin{tabular}{lcccccccccc}
\toprule
\multirow{2}{*}{Method}
& \multicolumn{2}{c}{\textbf{APPS Intro}}
& \multicolumn{2}{c}{\textbf{APPS Interview}}
& \multicolumn{2}{c}{\textbf{HumanEval}}
& \multicolumn{2}{c}{\textbf{MBPP}}
& \multicolumn{2}{c}{\textbf{Average}} \\
\cmidrule(lr){2-3} \cmidrule(lr){4-5} \cmidrule(lr){6-7} \cmidrule(lr){8-9} \cmidrule(lr){10-11}
 & AUROC & PRAUC & AUROC & PRAUC & AUROC & PRAUC & AUROC & PRAUC & AUROC & PRAUC \\
\midrule
\multicolumn{10}{l}{\textit{Qwen3-32B}} \\
\multicolumn{11}{l}{\textit{NL-derived baselines}} \\
\quad Mean Entropy & .450 & .561 & .532 & .291 & .490 & .868 & .544 & .617 & .504 & .584 \\
\quad Max Entropy & .460 & .559 & .491 & .283 & .523 & .889 & .532 & .632 & .502 & .591 \\
\quad Mean Probability & .469 & .602 & .520 & .316 & .530 & .900 & .580 & .658 & .525 & .619 \\
\quad Max Probability & .467 & .605 & .528 & .323 & .549 & .899 & .573 & .660 & .529 & .622 \\
\quad Consistency (BLEU) & .739 & .797 & .705 & .479 & .516 & .885 & .603 & .664 & .641 & .706 \\
\quad Consistency (VR) & \textbf{.799} & \underline{.843} & .762 & \underline{.587} & .574 & .898 & .663 & .713 & .700 & \underline{.760} \\
\quad Symb.\ Clustering & .665 & .776 & .610 & .477 & .654 & .922 & .560 & .660 & .622 & .709 \\
\multicolumn{11}{l}{\textit{Three axes (ours)}} \\
\quad Top-5 Entropy (LEX) & .573 & .644 & .616 & .369 & .613 & .902 & .592 & .657 & .599 & .643 \\
\quad Pseudo Consistency (ALGO) & .671 & .753 & .642 & .448 & .621 & .915 & .611 & .656 & .636 & .693 \\
\quad Generated Tests (FUNC) & .743 & .804 & \textbf{.770} & .571 & \textbf{.756} & \underline{.938} & \underline{.711} & \underline{.724} & \underline{.745} & .759 \\
\multicolumn{11}{l}{\textit{Ensemble (ours)}} \\
\quad Top-5 + Tests + Pseudo & \underline{.759} & \textbf{.884} & \underline{.768} & \textbf{.603} & \underline{.734} & \textbf{.944} & \textbf{.727} & \textbf{.738} & \textbf{.747} & \textbf{.792} \\
\midrule
\multicolumn{10}{l}{\textit{Qwen3-Coder-30B}} \\
\multicolumn{11}{l}{\textit{NL-derived baselines}} \\
\quad Mean Entropy & .670 & .710 & .655 & .458 & .581 & .957 & .585 & .627 & .623 & .688 \\
\quad Max Entropy & .780 & .806 & .722 & .546 & .678 & .977 & .570 & .607 & .688 & .734 \\
\quad Mean Probability & .612 & .659 & .619 & .518 & .662 & .951 & .510 & .610 & .601 & .684 \\
\quad Max Probability & .633 & .692 & .643 & .522 & .667 & .960 & .536 & .609 & .620 & .696 \\

\quad Consistency (BLEU) & .736 & .781 & .725 & .544 & .705 & \textbf{.981} & .600 & .619 & .692 & .731 \\
\quad Consistency (VR) & .689 & .733 & .667 & .483 & .578 & .962 & .528 & .560 & .616 & .685 \\
\quad Symb.\ Clustering & .612 & .697 & .550 & .434 & .517 & .960 & .592 & .616 & .568 & .677 \\
\multicolumn{11}{l}{\textit{Three axes (ours)}} \\
\quad Top-5 Entropy (LEX) & \textbf{.807} & .822 & \underline{.756} & .580 & .591 & .967 & .599 & .629 & .688 & .750 \\
\quad Pseudo Consistency (ALGO) & .721 & \underline{.838} & .610 & \textbf{.672} & .671 & .973 & .602 & .660 & .651 & \underline{.786} \\
\quad Generated Tests (FUNC) & .749 & .809 & .738 & .576 & \underline{.734} & .973 & \underline{.731} & \underline{.722} & \underline{.738} & .770 \\
\multicolumn{11}{l}{\textit{Ensemble (ours)}} \\
\quad Top-5 + Tests + Pseudo & \underline{.796} & \textbf{.852} & \textbf{.800} & \underline{.614} & \textbf{.746} & \underline{.980} & \textbf{.748} & \textbf{.767} & \textbf{.773} & \textbf{.803} \\
\bottomrule
\end{tabular}
\caption{Additional full uncertainty estimation results: Qwen3-32B and Qwen3-Coder-30B}
\label{tab:per_model_2}
\end{table*}

\end{document}